\pdfoutput=1

\documentclass[11pt]{article}

\usepackage[final]{acl}

\usepackage{times}
\usepackage{latexsym}
\usepackage{booktabs} 

\usepackage{multirow}
\usepackage{makecell}
\usepackage{float}
\usepackage{graphicx}

\usepackage[T1]{fontenc}

\usepackage[utf8]{inputenc}
\usepackage{microtype}

\usepackage{inconsolata}

\title{Fine-tuning Whisper on Low-Resource Languages for Real-World Applications}

\author{
  \textbf{Vincenzo Timmel\textsuperscript{1}},
    \textbf{Claudio Paonessa\textsuperscript{2}},
  \textbf{Manfred Vogel\textsuperscript{1}},
  \textbf{Daniel Perruchoud\textsuperscript{1}},
  \textbf{Reza Kakooe\textsuperscript{1}}
\\
\\
  \textsuperscript{1}University of Applied Sciences and Arts Northwestern Switzerland\\
  \textsuperscript{2}Noxenum.io\\
  \small{\{vincenzo.timmel, manfred.vogel, daniel.perruchoud, reza.kakooee\}@fhnw.ch, claudio.paonessa@noxenum.io}
}

\begin{document}

\maketitle
 
\begin{abstract}

This paper presents a new approach to fine-tuning OpenAI's Whisper model for low-resource languages by introducing a novel data generation method that converts sentence-level data into a long-form corpus, using Swiss German as a case study. Non-sentence-level data, which could improve the performance of long-form audio, is difficult to obtain and often restricted by copyright laws. Our method bridges this gap by transforming more accessible sentence-level data into a format that preserves the model’s ability to handle long-form audio and perform segmentation (by predicting timestamps) without requiring non-sentence-level data. Our data generation process improves performance in several real-world applications and leads to the development of a new state-of-the-art speech-to-text (STT) model for Swiss German. We compare our model with a non-fine-tuned Whisper and previous state-of-the-art Swiss German STT models, where our new model achieves higher BLEU scores. Our results also indicate that the proposed method is adaptable to other low-resource languages, supported by written guidance and code that allows the creation of fine-tuned Whisper models, which keep segmentation capabilities and allow the transcription of longer audio files using only sentence-level data with high quality.
\\
\end{abstract}

\section{Introduction}
Swiss German refers to the dialects spoken in the German-speaking regions of Switzerland. Due to the limited number of speakers, linguistic resources are scarce, also because Swiss German exists only as a spoken language, without any formal grammar or standardized written form. As a result, Swiss German STT systems are typically formulated as speech translation tasks, where Swiss German audio is transcribed into standard German text \cite{pluss-etal-2021-spc, pluss-etal-2022-sds, pluss-etal-2023-stt4sg}. Generally, "Swiss German transcription" or "Swiss German ASR" refers to converting spoken Swiss German directly into written Standard German, combining transcription and translation in a single step.

The Whisper models \cite{radford2022robust} developed by OpenAI are trained on a large-scale corpus of audio recordings and corresponding transcriptions obtained by web crawling. The dataset used for the multilingual version of Whisper includes samples from almost 100 different languages. After English and Chinese, German represents the third-largest part of the dataset, with 13'344 hours. The unexpectedly high transcription quality of Whisper for Swiss German audio and video (see Table~\ref{tab:Whispermerged}) and observed distinct hallucinations (discussed in Section \ref{sec:conclusions}) prove the presence of Swiss German audio in the original training dataset. While the model performs remarkably well for Swiss German, there is still considerable room for improvement for practical applications that require higher transcription quality, such as judicial interrogation transcripts or medical diagnosis and treatment orders.

In addition, fine-tuning solutions which fail to predict timestamps make it impossible to use Whisper for subtitling, multi-speaker conversation pattern analysis and other real-world applications\footnote{https://huggingface.co/blog/fine-tune-whisper}. Table \ref{tab:model_performance_sentence_dataset} illustrates how Whisper Large-v2 fine-tuned on sentences loses is capabilities to predict timestamps and starts to fail when the audio segment is getting longer and more difficult to predict, even though when evaluating it on the test-split of the sentence-level dataset, the fine-tuned model shows much improvement over the original Whisper Large-v2.

\begin{table}[h]
  \caption{Comparison of a rapidly spoken Swiss-German saying \cite{lingoda_schweizer_sprichwoerter}, transcribed by Whisper Large-v2. The model, fine-tuned on sentence-level data, fails to predict timestamps and performs worse than the original Large-v2. 
%
}
  \label{tab:model_performance_sentence_dataset}
  \centering
  \resizebox{0.9\linewidth}{!}{%
    \begin{tabular}{ l }
      \toprule
      \textbf{\footnotesize{Input Audio}} \\
      \footnotesize{...} \\
      \footnotesize{Ich zeig der, wo de Bartli de Moscht holt.} \\
      \footnotesize{...} \\
      \midrule
      \textbf{\footnotesize{Whisper Large-v2}} \\
      \footnotesize{...} \\
      \footnotesize{\texttt{[00:00:08]} Ich zeige dir, wo Bartli den Most holt. \texttt{[00:00:11]}} \\
      \footnotesize{...} \\
    \midrule
          \textbf{\footnotesize{Whisper Large-v2 (fine-tuned on sentences)}} \\
                \footnotesize{...} \\
      \footnotesize{Ich zeige dir, wo es die Bartli in den Most holt.} \\
      \footnotesize{...} \\
      \bottomrule
    \end{tabular}%
  }
\end{table}

In this paper, we focus on fine-tuning OpenAI's Whisper model for low-resource languages in real-world applications, using Swiss German as a case study. We evaluate the segmentation capabilities of Whisper after fine-tuning and demonstrate the beneficial effect of fine-tuning on long-form audios generated from sentence-level data. Finally, we evaluate the impact of the amount of training data on model performance when fine-tuning Whisper.

In particular, we address the following key research questions: 

\begin{itemize}
\item How can sentence-level datasets be adapted to effectively train Whisper models for longer audio sequences while maintaining segmentation and transcription quality?
\item How does fine-tuning Whisper affect its segmentation capabilities, especially when moving from sentence-level to long-form data? 
\item How does fine-tuning Whisper with additional datasets, such as pseudo-labeled long-form audio, affect its performance in various real-world scenarios?
\end{itemize}

By exploring these research questions, this paper provides insights into improving STT systems for low-resource languages through innovative data generation and fine-tuning strategies.

\section{Related work}
Despite the challenge of scarce data resources, recent advancements in speech translation for Swiss German have been substantial, driven in part by recent collections of high-quality sentence-level datasets from Swiss parliaments minutes and crowdsourcing initiatives such as SwissDial \cite{doganschönberger2021swissdial}, the Swiss Parliaments Corpus SPC \cite{pluss-etal-2021-spc}, SDS-200 \cite{pluss-etal-2022-sds}, and STT4SG-350 \cite{pluss-etal-2023-stt4sg}.

Prior to the release of Whisper, a commonly applied foundation model for building ASR systems was XLS-R \cite{DBLP:journals/corr/abs-2111-09296}. XLS-R is based on the wav2vec 2.0 architecture \cite{DBLP:journals/corr/abs-2006-11477} and was pre-trained on  436K hours of speech data in 128 languages. Previous research on Swiss German speech recognition has therefore often used the pre-trained XLS-R backbone \cite{pluss-etal-2022-sds, pluss-etal-2023-stt4sg, schraner2022swiss, paonessa-etal-2023-dialect}.

There are many papers and blog posts on the fine-tuning of Whisper \cite{gandhi2022fine, vaibhav2023fast, theo2023asr, openai2023peft, ma2023adapting, shamsian2024keyword, do2023using, ferraz2023multilingual, sicard2023spaicheextendingstateoftheartasr} and some focused on low-resource settings \cite{Pi_eiro_Mart_n_2024, hsu2024metawhisperspeechbasedmetaiclasr, liu2024, pillai2024multistagefinetuningstrategiesautomatic}. The problem of language forgetting is also discussed extensively, and it is shown that fine-tuning on a new language yields the best performance for the new language, but degrades the capabilities on existing languages \cite{qian24_interspeech}. However, it is rare to find papers that explicitly address the problem of fine-tuning Whisper for transcription of longer audios. Many papers also do not evaluate or discuss the segmentation capabilities after fine-tuning. Finally, it is uncommon to see evaluations of fine-tuned Whisper on out-of-distribution datasets. 

\section{Approach}

\subsection{Data Generation}
\label{sec:data_generation}

\begin{figure}[H]
    \centering
    \includegraphics[width=\linewidth]{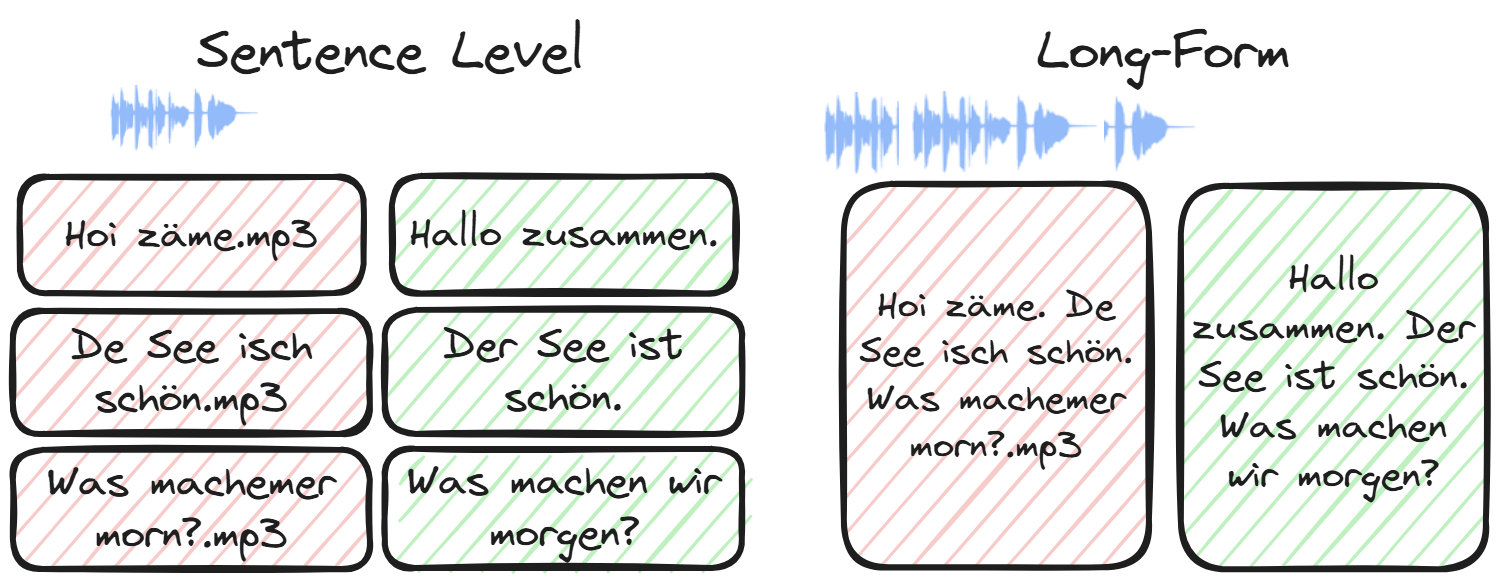}
    \caption{Illustration of generated long-form training data from sentence-level audios. Although timestamps are available via the length of the audio, they are not displayed here.}
    \label{fig:sentence_vs_long_form}
\end{figure}

The Whisper model works with a fixed input length of 30 seconds. Samples shorter than 30 seconds must be padded by appending zeros (silence). However, available datasets often consist of sentence-level samples, which are usually much shorter than 30 seconds. This is also true for Swiss German speech translation corpora such as SPC, SDS-200 and STT4SG-350. Using such training data to fine-tune Whisper is challenging because it requires significant padding for each sample. This carries the potential risk of compromising the model's ability to robustly handle long-form audio and predict timestamps, a crucial aspect of the model for many use cases. We therefore start with available sentence-level pairs of Swiss German audio and standard German transcriptions (see Table~\ref{tab:datasets}) and concatenate multiple sentences to synthetically generate long-form audios with corresponding segment timestamps, as shown in Figure \ref{fig:sentence_vs_long_form}.

The data generation strategy contains the following steps:
\begin{itemize}
\item \textbf{Timestamp Correction:} By using Voice Activity Detection (VAD), specifically leveraging Silero Models\footnote{\url{https://github.com/snakers4/silero-models}}, we correct the start and end timestamps of the resulting audio segments. 
\item \textbf{Noise Overlapping:} By simply concatenating two audio samples, the transitions often become noticeable because they abruptly change noise characteristics. To improve the transitions between consecutive samples, we employ a random overlapping technique that leverages the silence intervals detected by VAD at the beginning and end of each sample. By taking advantage of these silence parts, this enhancement accurately simulates consecutive audio segments. Together with Timestamp Correction, it also allows to create a speech overlap, such that two speakers speak over each other. 
\item \textbf{Speaker Retention:} For samples that include speaker identification, the probability of retaining the same speaker in successive samples is 50\%. This enhancement helps to create more realistic sequences in which speaker changes occur at a pace with natural speech patterns.
\\
\end{itemize}

\begin{figure}[htbp]
    \centering
    \includegraphics[width=\linewidth]{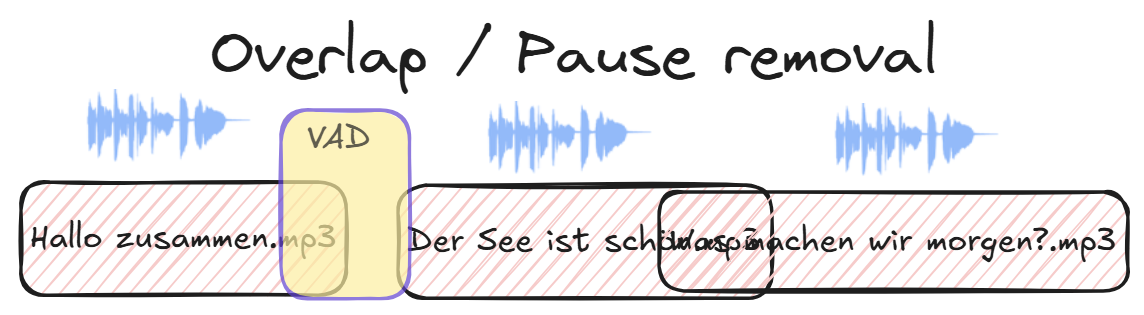}
    \caption{Illustration of the logical structure for stitching together sentences using VAD and overlap mechanisms. With the help of a VAD model, we precisely mark the start and end of speech. This allows us to vary the length of pauses between sentence and even introduce an overlap.}
    \label{fig:pause_overlap}
\end{figure}

Later in the study, in Table \ref{tab:evaluation_results_sentence_level_model}, we show the influence of these data generation strategies on different datasets. In Figure \ref{fig:pause_overlap} the general approach is shown. The beginning and the end of speech are detected and then, when concatenating sentences together, we can either: 
\begin{itemize}
\item do \textbf{Concat}, concatenate files as they are,
\item detect the end and start of speech and make the non-speech \textbf{Overlap} by up to 200ms when concatenating audios,
\item introduce a \textbf{Negative Overlap}, so that the speech of two sentences overlaps by 200ms.
\end{itemize}

\subsection{Training Details}

For model initialization, we use the Whisper Large-v2 weights, as initial tests showed it outperformed Whisper Large-v3 on the Swiss German datasets used. And we take advantage of its strong baseline performance for Swiss German by using the German language tag \textbf{DE} (see Section \ref{section:base}).

Using gradient check-pointing \cite{sohoni2022lowmemory}, gradient accumulation, and an 8-bit optimizer \cite{DBLP:journals/corr/abs-2110-02861}, we achieve an effective batch size of 256 on a single NVIDIA A100 40 GB GPU. Gradient check-pointing is applied to both the encoder and decoder, with 16 gradient accumulation steps and a per-step batch size of 16. Additionally, we apply stochastic depth \cite{huang2016deepnetworksstochasticdepth} to the encoder and decoder blocks. This setup results in a mixed-precision training run that takes about 42 hours.

We use a learning rate scheduler with a linearly increasing warm-up phase followed by a linear decay to zero, as described in the original Whisper training procedure \cite{radford2022robust}. During training, each sample has a 50\% chance of containing timestamps and a 50\% chance of containing prompts, mentioned in a comment on the OpenAI Whisper repository \footnote{\url{https://github.com/openai/Whisper/discussions/838}}.

\begin{table}[H]
  \caption{Training Hyperparameters}
  \label{tab:model_hyperparameters}
  \centering
  \begin{tabular}{ll}
    \toprule
    \textbf{Parameter}            & \textbf{Value} \\
    \midrule
    Optimizer                     & AdamW          \\
    Max. Learning Rate             & $2.0 \times 10^{-4}$ \\
    Weight Decay                  & 0.1            \\
    Warmup Updates                & 128            \\
    \midrule
    \multicolumn{2}{l}{\textbf{AdamW Specific Parameters}} \\
    $\beta_1$                     & 0.9            \\
    $\beta_2$                     & 0.98           \\
    $\epsilon$                    & $1.0 \times 10^{-9}$ \\
    \bottomrule
  \end{tabular}
\end{table}

Following the improved training procedure of Whisper Large-v2, we apply SpecAugment \cite{park19e_interspeech} during training with the same parameters as in \cite{radford2022robust}, summarized in Table \ref{tab:model_hyperparameters}.

\subsection{Train, Validation and Test Data}
\label{sec:train_val_test_data}
For our training data, we use the Swiss German sentence-level datasets \cite{pluss-etal-2021-spc, pluss-etal-2022-sds, pluss-etal-2023-stt4sg} with the predefined train and validation sets. For the train and validation split, unless mentioned otherwise, we use our data-generation pipeline explained in section \ref{sec:data_generation}. As additional training and validation data, we use Swiss Broadcasting Corporation (SRG) shows, pseudo-labeled (PL) by transcription with Whisper Large-v2. We selected 17 TV series, in which Swiss German is spoken.

As test data, we use the predefined split of the Swiss German sentence-level datasets (not processed by our pipeline and thus stay as single sentences) and a Dataset-A containing a manually transcribed doctor-patient conversation obtained from a confidential phone call. Due to data privacy, this dataset cannot be disclosed and remains a closed source dataset. As an additional test set, we use SRG data from 5 TV series for which manual transcriptions are available, i.e.: \textit{Einstein}, \textit{Puls}, \textit{Impact Investigativ}, \textit{SRF Kids News}, and \textit{SRF ohne Limit}. In contrast to the pseudo-labeled SRG  train and validation data, we use as test set subtitles manually created by SWISS TXT, a subsidiary of SRG.

Because we have reasonable suspicion (see Section \ref{sec:conclusions}) 
that OpenAI has data from the SRG in its Whisper training corpus, we only considered SRG data for the validation and test set broadcasted after the release of Whisper Large-v2 to allow a fair comparison with our baseline, the Whisper Large-v2 base model.

The total hours of data used for training, validation and testing are presented in Table \ref{tab:datasets}.

\begin{table}[H]
  \caption{Overview of the datasets used for training, validation, and testing, including totals per split.}
  \label{tab:datasets}
  \centering
  \begin{tabular}{lccc}
    \toprule
    \textbf{Name (Variant)} & \textbf{Split} & \textbf{Hours} & \textbf{\# Speakers} \\
    \midrule
    SDS-200 (Clean) & Train & 50 & 1,799 \\
    STT4SG-350 (All) & Train & 276 & 219 \\
    SPC (0.9 IOU) & Train & 176 & 194 \\
    SRG (PL) & Train & 406 & -- \\
    \midrule
    Total & & 908 & $>$ 2,212 \\
    \midrule
    SDS-200 (Clean) & Val & 5.2 & 288 \\
    STT4SG-350 (All) & Val & 21 & 219 \\
    SRG (PL) & Val & 20 & -- \\
    \midrule
    Total & & 46.2 & $>$ 507 \\
    \midrule
    SDS-200 (Clean) & Test & 5.2 & 281 \\
    STT4SG-350 (All) & Test & 34 & 56 \\
    SPC (0.9 IOU) & Test & 6 & 26 \\
    SRG (SWISS TXT) & Test & 20 & -- \\
    Dataset-A & Test & 0.22 & 2 \\
    \midrule
    Total & & 65.42 & $>$ 365 \\
    \bottomrule
  \end{tabular}
\end{table}

\begin{figure}[H]
    \centering
    \includegraphics[width=\linewidth]{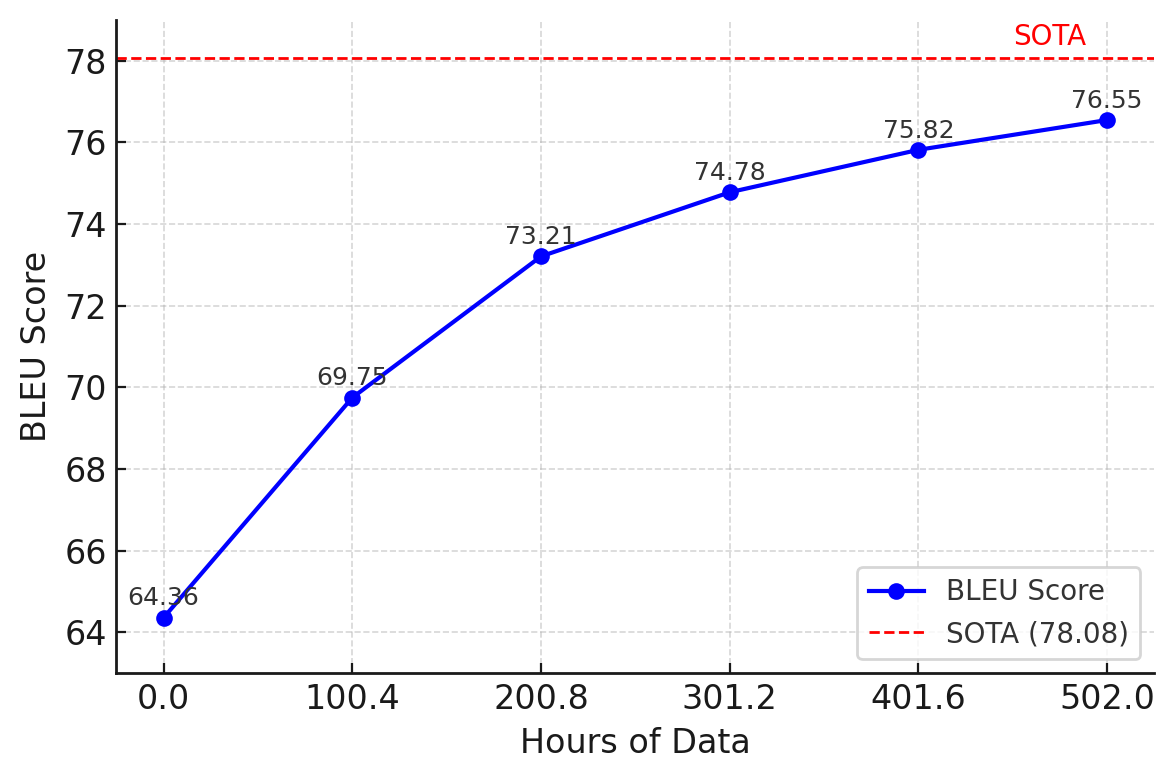}
    \caption{
    BLEU score on the STT4SG-350 test set vs. amount of training data (given in Table \ref{tab:datasets}) used for fine-tuning. The model evaluated at 0 hours of training data corresponds to the original Whisper Large-v2. The SOTA model is discussed in section \ref{sec:overcome_shortcomings}.}
    \label{fig:bleu_vs_hours_of_data_sota}
\end{figure}

In Figure \ref{fig:bleu_vs_hours_of_data_sota}, we analyze the relationship between the BLEU score \cite{papineni-etal-2002-bleu} on the STT4SG-350 test set and the amount of training data used for fine-tuning. For training, we used the 502 hours long-form corpus consisting of SDS-200, STT4SG-350 and SPC, but we do not include the pseudo-labeled data to show what can be expected from high-quality labeled data. The models were trained using hierarchically nested datasets, each partition holding approximately 20\% of the training data. Training was run until the word error rate (WER) on the validation set showed no more improvement. Once the training had stabilized, the best performing model on WER was selected.

Based on the unabated rise of the curve in Figure \ref{fig:bleu_vs_hours_of_data_sota}, it's reasonable to deduct that more training data will improve model performance further.

\section{Experiments}
\subsection{Evaluation}
\label{sec:evaluation}
In all experiments, the computed metrics are derived from the WhisperX\footnote{\url{https://github.com/m-bain/whisperX}} versions of the models using fp16 precision. The beam size is configured to 5, \textbf{VAD} is enabled and the language tag is \textbf{DE}. For the sentence-level datasets, we report the WER or the BLEU metric or both, if the layout allows it.
For the long-form test dataset we replace the BLEU metric with SubER \cite{wilken2022suber}; this allows us to incorporate a metric that also assesses the quality of segmentation by timestamp prediction. Before calculating the metrics, all sentences are transformed to lowercase, and punctuation is removed. For the BLEU metric we specifically use sacreBLEU \cite{post-2018-call} with default parameters.

\subsection{Base Results}
\label{section:base}

In Table~\ref{tab:Whispermerged} we compare the performance of the original Whisper Large models (v2 and v3) on our datasets. Interestingly, Whisper Large-v3 exhibits an improvement over its predecessor, Large-v2, only on the STT4SG-350 test set. It is noteworthy that the 24,605 samples in the STT4SG-350 test set yielded identical transcripts for both models in 11,340 instances (46.1\%). Conversely, both the SPC and SDS-200 test sets reveal a slight decline in performance for Whisper Large-v3. The gap between the two models is most evident on the SRG dataset, where the Large-v2 model yields noticeably better performance.

\begin{table}[H]
  \centering
  \small 
  \caption{Performance of the original Whisper models on various sentence-level test sets. SubER is shown on the long-form SRG data.}
  \label{tab:Whispermerged}
  \begin{tabular}{llrrr}
    \toprule
    \textbf{Test Dataset} & \textbf{Model} & \textbf{WER} & \textbf{BLEU} & \textbf{SubER} \\
    \midrule
    \multirow{2}{*}{SPC} & Large-v2 & \textbf{28.21} & \textbf{58.08} & - \\
                        & Large-v3 & 28.94 & 57.90 & - \\
    \midrule
    \multirow{2}{*}{SDS-200} & Large-v2 & \textbf{27.69} & \textbf{57.35} & - \\
                             & Large-v3 & 27.88 & 57.00 & - \\
    \midrule
    \multirow{2}{*}{STT4SG-350} & Large-v2 & 22.41 & \textbf{64.15} & - \\
                                & Large-v3 & \textbf{22.01} & 64.13 & - \\
    \midrule
    \multirow{2}{*}{SRG (SWISS TXT)} & Large-v2 & \textbf{28.42} & \textbf{63.61} & \textbf{30.63} \\
                           & Large-v3 & 38.69 & 56.31 & 42.58 \\
    \bottomrule
  \end{tabular}
\end{table}

Based on this comparison, there is no indication of a noticeable improvement on Swiss German audio when using Whisper Large-v3 instead of Large-v2.

\subsection{Segmentation Forgetting \& Out-of-Distribution Performance Degradation}
\label{subsec:seg_forg}

To assess the impact of fine-tuning without timestamps on both segmentation capabilities and performance on out-of-distribution data, we conducted experiments comparing models trained on different data generation parameters with our pipeline.  Table~\ref{tab:evaluation_results_sentence_level_model} presents the BLEU and SubER metrics for each model across various test datasets.
The following parameters for the data generation are compared:

\begin{itemize}
  \item \textbf{Sentence-level}: Training is conducted at the sentence level with padding.
  
  \item \textbf{Concat}: No adjustments are made; sentences are simply concatenated together.
  
  \item \textbf{Speaker\_ret}: There is a 50\% chance that two consecutive sentences originate from the same speaker without reusing any sentences.
  
  \item \textbf{Overlap}: There is a 50\% chance that two sentences overlap, meaning the speech of the sentence begins immediately as the speech of the preceding sentence ends.
  
  \item \textbf{Neg\_overlap}: There is a 10\% chance that when audios overlap, the speech of two separate sentences overlaps by 200ms, simulating scenarios where two speakers talk simultaneously.
  
  \item \textbf{All}: This method combines all the aforementioned preprocessing techniques.
  
\end{itemize}

\begin{table}[H]
  \caption{
   Comparison of BLEU and SubER metrics for Whisper Large-v2, fine-tuned on sentence-level (first row) and differently generated long-form data (Large-v2 refers to the original model without fine-tuning). The results on SRG and Dataset-A show that the sentence-level fine-tuning performs noticeably worse compared to training on long-form data.
  }
  \label{tab:evaluation_results_sentence_level_model}
  \centering
    \small 
  \begin{tabular}{llcc}
    \toprule
    \textbf{Test Dataset} & \textbf{Data/Model} & \textbf{BLEU} & \textbf{SubER} \\
    \midrule
    \multirow{7}{*}{STT4SG-350} 
      & \footnotesize{Sentence-level} & \textbf{77.38} & - \\
      & \footnotesize{Concat}        & 76.08 & - \\
      & \footnotesize{Speaker\_ret}   & 76.91 & - \\
      & \footnotesize{Overlap}        & 72.12 & - \\
      & \footnotesize{Neg\_overlap}   & 76.53 & - \\
      & \footnotesize{All}            & 76.55 & - \\
      & \footnotesize{Large-v2}       & 64.15 & - \\
    \midrule
    \multirow{7}{*}{SRG (SWISS TXT)} 
      & \footnotesize{Sentence-level} & 47.57 & 51.34 \\
      & \footnotesize{Concat}        & 51.34 & 42.87 \\
      & \footnotesize{Overlap}        & 49.68 & 44.15 \\
      & \footnotesize{Speaker\_ret}   & 52.51 & 41.07 \\
      & \footnotesize{Neg\_overlap}   & 50.63 & 43.56 \\
      & \footnotesize{All}            & 51.62 & 41.64 \\
      & \footnotesize{Large-v2}       & \textbf{63.61} &\textbf{30.63} \\
    \midrule
    \multirow{7}{*}{Dataset-A} 
      & \footnotesize{Sentence-level} & 35.01 & 55.44 \\
      & \footnotesize{Concat}        & 46.80 & 41.94 \\
      & \footnotesize{Speaker\_ret}   & 46.79 & 41.99 \\
      & \footnotesize{Overlap}        & 45.84 & 43.62 \\
      & \footnotesize{Neg\_overlap}   & 45.40 & 43.03 \\
      & \footnotesize{All}            & 47.22 & 41.15 \\
      & \footnotesize{Large-v2}       & \textbf{48.89} & \textbf{39.12} \\
    \bottomrule
  \end{tabular}
\end{table}

The model fine-tuned solely on padded sentence-level samples exhibits a substantial decline in timestamp prediction accuracy, evidenced by a SubER score exceeding 51 on the SRG dataset, even with the help of VAD, which substitutes timestamps. Because we use WhisperX to evaluate the models, the timestamps are given by the VAD-Model. Without WhisperX, the metrics of the sentence-level Model would be much worse.

While the sentence-level model attains with 77.38 the highest BLEU score on the STT4SG-350 dataset, it performs poorly on longer audio sequences in SRG and Dataset-A, especially when the segmentation quality is taken into account. This underscores its limitations with out-of-distribution data. In comparison, models trained with the generated long-form dataset — especially 'All' — demonstrate better generalization, maintaining higher BLEU scores and lower SubER scores across different datasets, nearly reaching the original Whisper Large-v2 on the Dataset-A on BLEU. 

The original Large-v2 model without fine-tuning, outperforms the sentence-level model on long-form datasets. This suggests that fine-tuning exclusively on sentence-level, which was data processed into long-form, degrades the performance on unseen datasets. Incorporating long-form audio and datasets with diverse distribution into the training process is essential for preserving segmentation capabilities and ensuring robust performance across diverse data distributions. 

\subsection{Overcome Shortcomings}
\label{sec:overcome_shortcomings}

Despite the measures taken to simulate long-form audio, our fine-tuning procedure leads to a reduction in segmentation and transcription quality when applied to real long-form audio, as shown  by the SubER metric given in Table~\ref{tab:evaluation_results_sentence_level_model} for the Dataset-A and SRG datasets. To address this issue, we enrich the training dataset by incorporating samples from the specific distribution of the intended prediction target, in our case, pseudo-labeled SRG data mentioned in Section \ref{tab:datasets}.

As part of our methodology to mitigate language forgetting in the final model training, we use the Mozilla Common Voice 13 German dataset \cite{commonvoice:2020}. Using the data generation pipeline described in Section \ref{sec:data_generation}, we curated a subset of 15,000 samples, each lasting 30 seconds, resulting in 125 hours of additional training data. The train, validation, and test set splits were taken as defined by Mozilla Common Voice Version 13.

This leads to a model fine-tuned on long-form audio with our generated corpus based on the three sentence-level datasets (SPC, SDS-200, and STT4SG-350), the pseudo-labeled dataset SRG (PL) based on Swiss German TV shows, and the German part of the Common Voice 13 training data, concatenated as described above. This strategy significantly improves the model performance, as shown in Table~\ref{tab:evaluation_results_sota_model}, and leads to a new state-of-the-art model for Swiss German speech-to-text, referred to as {\bf SOTA}.

\begin{table}[h]
  \caption{Our new Whisper Large-v2 based SOTA model, fine-tuned on long-form audio created with data generation method 'All' and supplemented with SRG (PL) and Common Voice 13 de, compared to the original Whisper Large-v2}
  \label{tab:evaluation_results_sota_model}
  \centering
    \small 
  \begin{tabular}{ llrrr }
    \toprule
    \textbf{Test Dataset}  & \textbf{Model}  & \textbf{WER}  & \textbf{SubER}  & \textbf{BLEU} \\
    \midrule
    \multirow{2}{*}{SPC}    & Our SOTA       & \textbf{20.98} & -         & \textbf{68.34} \\
                            & Large-v2       & 28.21         & -         & 58.08 \\
    \midrule
    \multirow{2}{*}{SDS-200}& Our SOTA       & \textbf{16.70} & -         & \textbf{72.69} \\
                            & Large-v2       & 27.92          & -         & 57.00 \\
    \midrule
    \multirow{2}{*}{STT4SG-350} & Our SOTA   & \textbf{12.11} & -         & \textbf{78.08} \\
                                & Large-v2   & 22.41          & -         & 64.13 \\
    \midrule
    \multirow{2}{*}{SRG (SW-TXT)}  & Our SOTA  & \textbf{26.31} & \textbf{29.76} & \textbf{64.67} \\
                            & Large-v2       & 28.42          & 30.63     & 63.61 \\
    \midrule
    \multirow{2}{*}{Dataset-A} & Our SOTA    & \textbf{34.50} & \textbf{35.31} & \textbf{51.40} \\
                               & Large-v2    & 38.00         & 39.12     & 48.89 \\
    \midrule
    \multirow{2}{*}{CV13 de}& Our SOTA       & \textbf{6.42}  & -         & - \\
                            & Large-v2       & 6.53           & -         & - \\

    \bottomrule
  \end{tabular}
\end{table}

\subsection{Dialect Comparison}

Since the STT4SG-350 test dataset contains identical sentences in 7 different dialects, it allows a fair comparison of model performance in terms of dialect-specific accuracy.

The results in Table~\ref{tab:wer_scores_compact} show large differences in the performance of the original Large-v2 model across the different dialect regions. In contrast, our fine-tuned SOTA model exhibits improved WER over a much narrower range across all dialects. For the reader, the improvements over other Swiss-German ASR models based on Wav2Vec (XLS-R) and Transformer (TF) architectures \cite{schraner2022swiss} are also shown.

\begin{table}[h]
  \caption{WER for Swiss German dialects on the STT4SG-350 test set for selected models; XLS-R and TF are older models based on Wav2Vec (XLS-R) and Transformer (TF) architectures  \cite{schraner2022swiss}.}
  \label{tab:wer_scores_compact}
  \centering
  \begin{tabular}{ lcccc }
    \toprule
    \textbf{Dialect} & \textbf{Large-v2} & \textbf{SOTA} & \textbf{XLS-R} & \textbf{TF} \\
    \midrule
    BS & 25.02 & \textbf{12.72} & 16.30 & 21.24 \\
    BE & 25.92 & \textbf{13.68} & 15.74 & 20.96 \\
    GR & 19.59 & \textbf{11.45} & 14.32 & 17.29 \\
    IS & 17.63 & \textbf{10.73} & 13.26 & 16.37 \\
    OS & 21.27 & \textbf{12.45} & 16.45 & 18.58 \\
    VS & 29.31 & \textbf{12.72} & 17.75 & 22.64 \\
    ZH & 18.29 & \textbf{11.03} & 13.41 & 17.30 \\
    \bottomrule
  \end{tabular}
\end{table}

\section{Conclusions}
\label{sec:conclusions}

A key advantage of OpenAI’s Whisper model is its ability to process audio of arbitrary length with built-in segmentation capabilities. However, fine-tuning such a model on sentence-level datasets while preserving these features is a significant challenge.

This paper demonstrates the potential of fine-tuning Whisper for low-resource languages, using Swiss German as a case study, and addresses the three research questions posed in the introduction. First, the paper shows how sentence-level datasets can be effectively adapted for training on longer audio sequences through a novel data generation pipeline, including techniques such as timestamp correction, noise overlapping, and speaker retention. These methods enable the generation of realistic long-form audio data that preserves segmentation and transcription quality.

Second, the fine-tuning approach significantly improves the model’s segmentation capabilities, particularly for long-form data, compared to sentence-level models. By evaluating the segmentation performance with SubER metrics, the study highlights the benefits of incorporating diverse training data and demonstrates improved robustness for timestamp prediction and audio segmentation.

Finally, the inclusion of additional datasets, such as pseudo-labeled long-form audio from Swiss Broadcasting Corporation shows, improves the model's performance in real-world applications.  We also show how to maintain performance in other languages by supplementing the training data with samples from those languages, thereby mitigating catastrophic forgetting. The results show that this method generalizes well to out-of-distribution datasets, achieving state-of-the-art performance in Swiss German speech-to-text tasks and suggesting broader applicability to other low-resource languages. 

In addition, we have highlighted that while the model may improve on data from the same distribution as the training data, in reality the model preforms worse on out-of-distribution data, as shown in table 5. This underscores the importance of creating or acquiring evaluation datasets that closely mimic the intended deployment environment, ensuring the ASR system’s robustness and usefulness in real-world applications.

Our research lays the groundwork for future work on data preparation and fine-tuning for OpenAI's Whisper model, especially in low-resource settings. 
For this we provide a simple framework, addressing catastrophic forgetting through long-form data generation and pseudo-labeling, enabling robust transcription even with limited datasets. The code for our data generation procedure \footnote{\url{https://github.com/i4ds/Whisper-prep}} and model fine-tuning \footnote{\url{https://github.com/i4ds/Whisper-finetune}} is publicly available.

Additionally, we observed distinct hallucinations of Whisper Large-v2 mentioning Swiss subtitling companies, such as being able to reliably trigger Whisper to transcribe "Untertitel von SWISS TXT" - a watermark of SWISS TXT that is only present in the subtitle files, never in the audio - when asked to transcribe the title music of the SRF Meteo show or when music is being played in "SRF bi de Lüt".

\section{Future work}
As we have extensively analyzed and evaluated different methods to generate long-form data from sentence-level data, the combination of the data generation and methods to avoid catastrophic forgetting, as presented by \cite{qian24_interspeech} by using Elastic Weight Consolidation \cite{kirkpatrick2017overcoming}, could be a next research topic. Another potential next step involves diversifying the data sources by augmenting the pseudo-labeled datasets with additional real-world data, including a broader range of TV programs, varied conversational contexts, and noisy environments. This expansion aims to enhance the robustness and generalization capabilities of the models. Notably, preliminary experiments indicate that a fine-tuned Whisper Large-v3 model performs particularly well on conversational speech, highlighting its potential superiority in this context and emphasizing the need for a large corpus of freely spoken dialogues.

\bibliography{custom}

\end{document}